\documentclass{INTERSPEECH2023}

\interspeechcameraready 
\usepackage{booktabs}
\usepackage{amsmath}
	
\title{Improving End-to-End SLU performance with Prosodic Attention and Distillation}
\name{Shangeth Rajaa}
\address{
Skit.ai\\
Bengaluru, India}
\email{shangeth.rajaa@skit.ai}

\begin{document}

\maketitle
 
\begin{abstract}
Most End-to-End SLU methods depend on the pretrained ASR or language model features for intent prediction. However, other essential information in speech, such as prosody, is often ignored. Recent research has shown improved results in classifying dialogue acts by incorporating prosodic information. The margins of improvement in these methods are minimal as the neural models ignore prosodic features. In this work, we propose prosody-attention, which uses the prosodic features differently to generate attention maps across time frames of the utterance. Then we propose prosody-distillation to explicitly learn the prosodic information in the acoustic encoder rather than concatenating the implicit prosodic features. Both the proposed methods improve the baseline results, and the prosody-distillation method gives an intent classification accuracy improvement of 8\% and 2\% on SLURP and STOP datasets over the prosody baseline.

\end{abstract}


\section{Introduction}

Natural Language Understanding (NLU) is one of the critical components of many conversational AI systems for interpreting and extracting meanings from user input. Intent classification is a common NLU task involving identifying a user's intention behind their utterance. Most existing systems use a two-stage pipeline approach for intent classification. Automatic speech recognition (ASR) is used to transcribe the spoken utterance, and an NLU model is then used to classify the intent. However, these pipeline intent classification approaches have a few drawbacks. These methods are prone to error propagation due to ASR transcript errors \cite{tran2020neural}, which can adversely affect the intent prediction performance of the NLU model. This two-stage pipeline approach increases the computation requirement and latency of the intent prediction systems. 

 Some important information in the speech signal, such as prosody(pitch, tempo, speaking rate, etc.) and speaker information(speaker's accent, gender, etc.), are lost after ASR. The NLU models only use the text transcript of the utterance for intent classification. This ASR+NLU pipeline method is based on the hypothesis that only the semantic meaning of the utterance is required for intent classification. In contrast, humans often incorporate a speaker's prosody to understand the utterance's intention.

 Recent works in intent classification have used an end-to-end spoken language understanding(SLU) approach to identify the intent directly from the speech signal. SLU models usually have fewer parameters and are much faster during deployment. SLU models could take advantage of other aspects of speech, such as prosody, and speaker information, unlike the NLU pipeline method. Most works in end-to-end SLU focus on ASR pretraining \cite{chen2018spoken}, or joint training \cite{lugosch2019speech} for intent classification. \cite{jiang2021knowledge}, \cite{sunder2022tokenwise} and \cite{huang2020leveraging} attempt to model the representations of pretrained text encoder such as BERT for the end-to-end SLU model. Few works modeled the acoustic encoder with a pretrained ASR model and used the utterance's semantic or phonetic information for intent classification. Very few methods take advantage of other aspects of speech, such as prosodic information in the speech signal for Speech to Intent(S2I).

\begin{figure}[]
  \includegraphics[width=\linewidth]{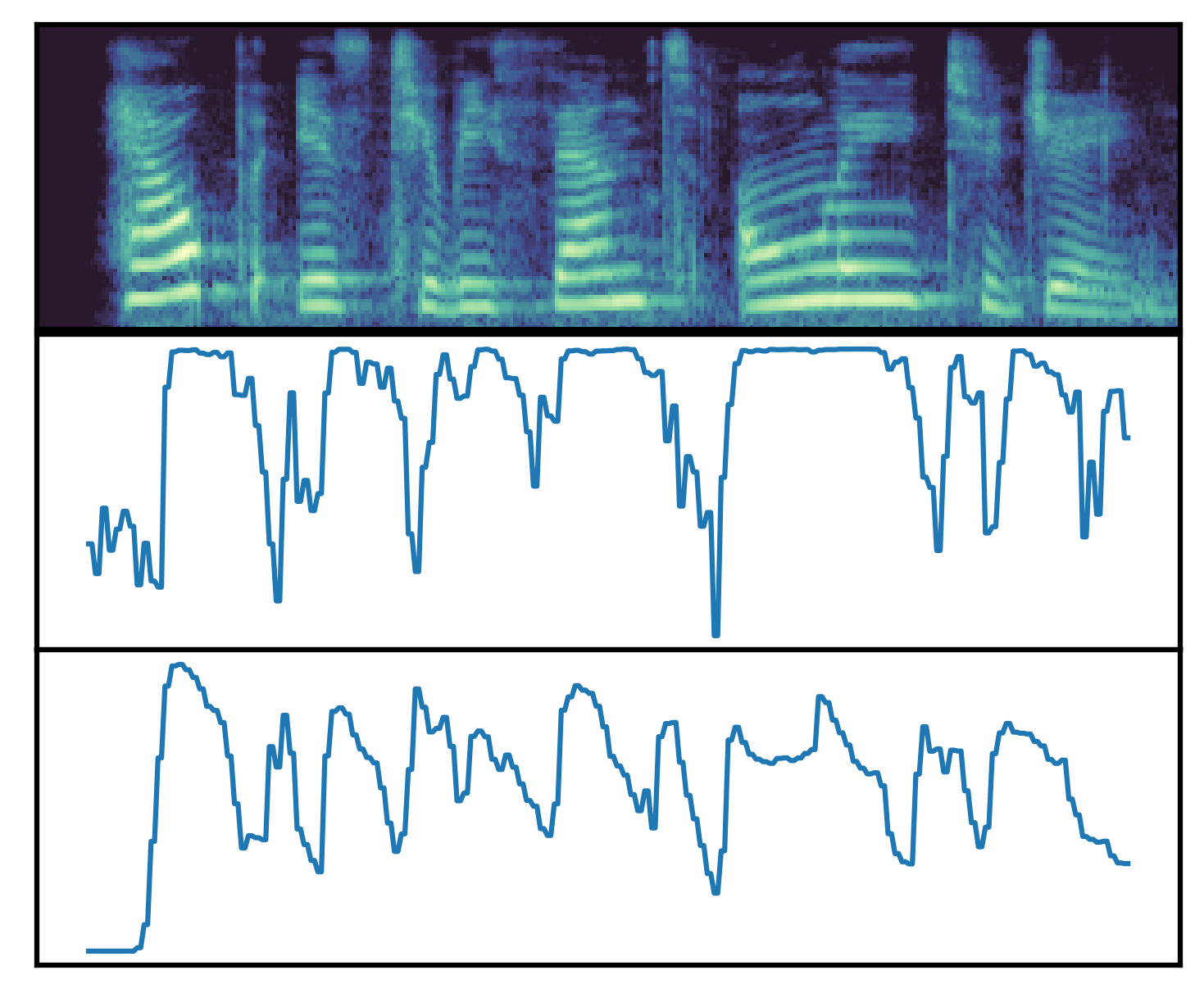}
  \caption{An example of Log Mel Spectrogram(upper), Log pitch(middle), and Total energy(lower) of a speech utterance from the SLURP dataset.}
  \label{fig:prosody_features}
  \vspace{-4mm}
\end{figure}


Prosody is fundamental in human speech communication \cite{wallbridge2021s}, capturing information beyond the linguistic and semantic meaning in an utterance\cite{dahan2015prosody}. Prosody helps disambiguate meaning, dialects, intents, sentiment, and other communication aspects not reflected in the text transcripts \cite{tran2020neural}. The stress, pitch, intonation, and timing pattern of an utterance can convey the speaker's intention("I have to go now." vs. "I have to go NOW?". The stress at the word "now" changes the intent/act of the utterance from statement to surprise or question). \cite{https://doi.org/10.48550/arxiv.2212.13015} shows that self-supervised pretrained(SSL) speech encoders perform better in S2I than ASR pretrained encoders, as SSL representations also contain prosodic information. \cite{tran2020neural} uses convolutional neural networks to encode the prosodic features such as pitch and energy to improve the performance of text-to-intent NLU models. Recently, \cite{wei2022neural} proposed a neural prosody encoder for end-to-end dialogue act classification. However, the margins of improvement in the evaluation metrics of these methods were minimal. They require prosodic features to be computed and encoded during inference, which may increase the prediction latency.

We hypothesize that prosodic features can help intent classification in two ways: Firstly, some words in the utterance are more critical than others in identifying the intent, and prosodic features can help identify these important words or provide an attention map. Secondly, the direction of change in prosodic features, for example, the slope of the pitch contour, can change the intention of any particular word in the utterance. Therefore, although the semantic meaning or ASR features of the words in the utterance will contribute more to intent prediction, the gap between current S2I models and humans in intent prediction is the prosodic information. Normal concatenation of prosodic and semantic/phonetic features may not fill this gap. As semantic/phonetic features contribute more to intent classification, the neural model tends to weigh those more and ignore prosodic features. From our experiment of training an SLU model by concatenating prosodic and ASR features, we found that the sum of model weights for the ASR features was five times higher than prosodic features. As a result, the model ignores prosodic features when classifying intent. So we need to incorporate prosodic features better to get the full advantage of prosodic information for intent classification.

\textbf{Our Contribution:} In this work, we first show that prosodic features can be used as an attention map to find which part of the utterance contributes more to the speaker's intention. Using prosody-based attention(prosody-attention) with the baselines improves the evaluation metrics. Then we propose prosody-distillation to learn the explicit prosodic information in the acoustic encoder without the need for explicit prosodic features with the help of a teacher prosody model. We perform knowledge distillation in two ways: prosody attention distillation and prosody feature distillation. Using both distillation methods improves the performance of the intent classification by huge margins on two public SLU datasets. We also perform a few ablation studies to find the impact of different hyperparameters or choices of methods on the performance of the prosody-distillation method.  Finally, we visualize the attention maps of trained prosody-attention and prosody-distillation models to understand which part of the utterance these models attend to for intent classification.


\section{Proposed Method}

\subsection{Prosodic Features}
\label{ssec:prosodic-feature}
 For the speech signal $X$, prosodic features $P = \{P_1, P_2, ... P_T\}$ with $T$ frames are computed. This work uses two basic types of prosodic features: pitch and energy.

\begin{itemize}
  \item Pitch: We extract the Log Pitch $(p^1)$,  Normalized Cross Correlation Function (NCCF)$(p^2)$, and derivative of Pitch $(p^3)$ for the speech signal using \cite{bernard2023shennong}. 
  \item Energy: We compute the total energy $(e^1)$, the energy of the upper 40 mel frequency bands$(e^2)$, and the energy of the lower 40 mel frequency bands$(e^3)$ using the 80-channel log Mel spectrogram computed on 25 millisecond windows with a stride of 10 milliseconds from the speech signal.
\end{itemize}

In total, we use six prosodic features $P_t = (p^1_t, p^2_t, p^3_t, e^1_t, e^2_t, e^3_t)$ for $t=T$ frame. Figure \ref{fig:prosody_features} shows an example of the Log Mel Spectrogram of the speech signal, the Log Pitch of the speech signal $(p^1)$, and the Total Energy of Mel Spectrogram $(e^1)$.

\begin{figure}
\centering
  \includegraphics[width=\linewidth]{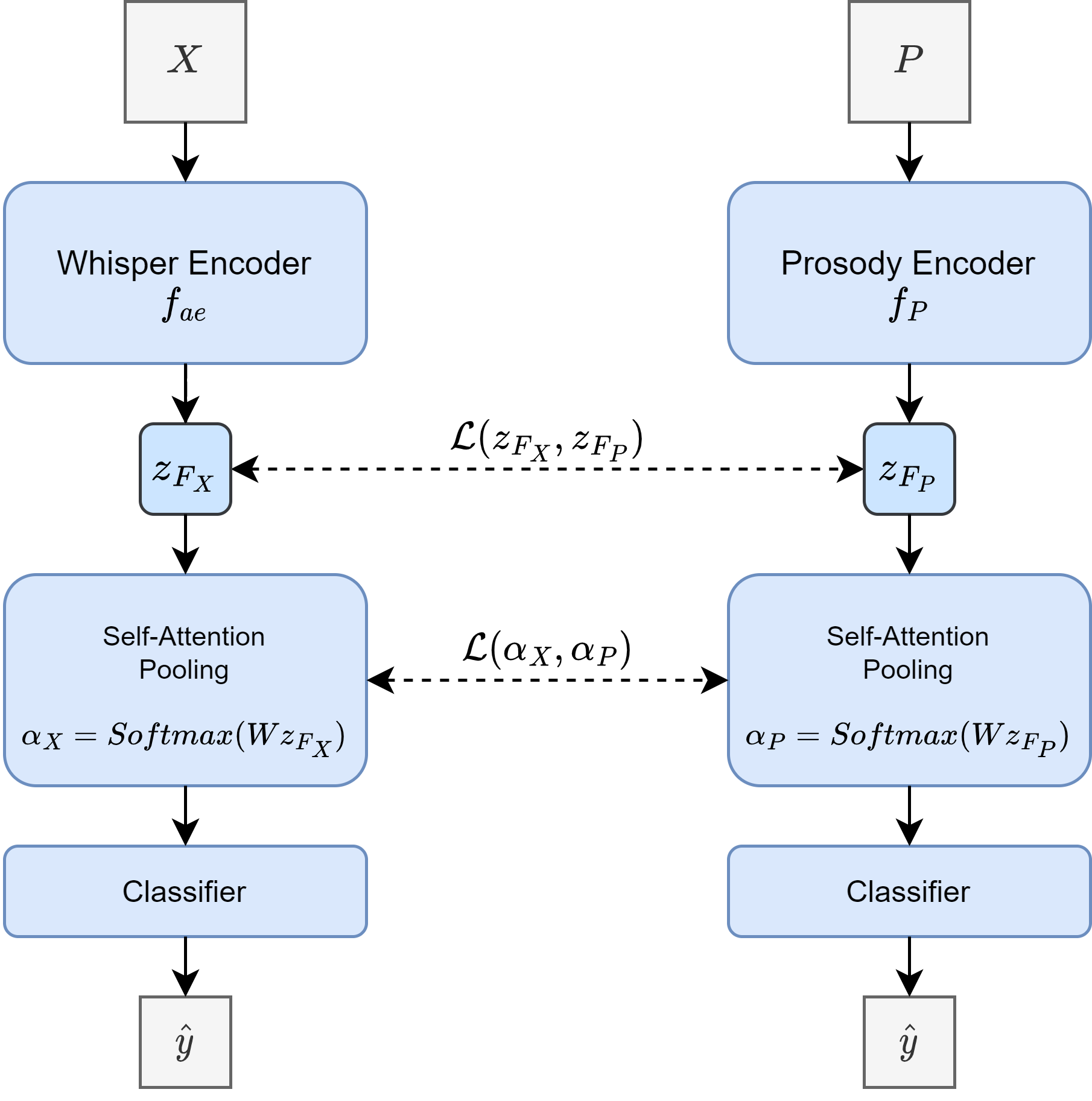}
  \caption{The proposed prosody-distillation method}
  \label{fig:prosody_attention}
  \vspace{-5mm}
\end{figure}

\subsection{Prosodic Attention}
Given the Speech signal $X$, the acoustic encoder $f_{ae}$ encodes the speech signal into frame-level representation $z_F$ with $T$ time frames. The frame-level representations $z_F$ are pooled into a single utterance level representation $z_U$ using a modified Self-Attention Pooling(SAP) \cite{safari2020self}. Instead of using $z_F$ to generate the attention weights $\alpha$, we generate the weights using the prosodic features $P$, and $z_U$ is calculated as the weighted sum of $z_F$ with weights $\alpha$ and is given by Equation \ref{eqn:self-attn}. The utterance level representation $z_U$ is passed through a linear classifier for intent prediction, and classification Cross-Entropy(CE) loss is used to train the model.

\begin{equation} 
\label{eqn:self-attn}
\begin{aligned}
z_F = f_{ae}(X) \\
\alpha = Softmax(W P) \\
z_U = \sum{\alpha z_F}
 \end{aligned}
\end{equation}

where $W$ is the learnable parameter of self-attention pooling layer and $\alpha = \{\alpha_1, \alpha_2, ..., \alpha_{T}\}$ are the attention weights for each frame of $z_F$. 

\subsection{Prosody Distillation}
Prosody-distillation method aims to learn the explicit prosodic information without the need for implicit prosodic features for intent classification. 
A teacher prosody model is pre-trained to recognize intent from only an utterance's prosodic features $P$. The teacher prosody model has a prosody encoder $f_P$, followed by self-attention and a classifier layer. The encoded representation of the prosody encoder $f_P$ will be rich in prosodic information that helps in predicting intents, and the self-attention layer of the teacher model can identify the frames in the utterance that captures the speaker's intent.

Then we train a student intent classification model, which learns the attention map and encoded prosodic features from the teacher prosody model. A multi-task learning (MTL) setting is used to train the model to classify the intents with the speech signal and learn prosodic information from the teacher model. During inference, only the student model is used for intent classification.

The student model has an acoustic encoder $f_{ae}$, a self-attention layer, and a linear layer for classification. The student model is trained with two forms of distillation from the teacher model, attention distillation, where the SAP layer of the student model learns to attend to time frames similar to prosody-attention, and feature distillation, where $f_{ae}$ learns to encode prosodic information which is necessary for intent classification. Mean Squared Error(MSE) is used as the loss for the distillation components. The attention distillation loss is calculated between the attention weights of SAP layers of the student and teacher models. The feature distillation loss is calculated between the frame-level feature maps of $f_{ae}$ and $f_P$. The final distillation loss  $\mathcal{L}_{dis}$ is a sum of attention and feature distillation loss. For intent classification, the student model is also trained using CE loss $\mathcal{L}_{cls}$. The total loss $\mathcal{L}_{total}$ is a weighted sum of both the distillation loss $\mathcal{L}_{dis}$ and classification loss $\mathcal{L}_{cls}$ and is given by Equation \ref{eqn:prosody-distillation}.

\begin{equation} 
\centering
\label{eqn:prosody-distillation}
\begin{aligned}
\mathcal{L}_{cls} = CE(y, \hat{y}) \\
\mathcal{L}_{dis} = MSE(z_{F_P}, z_{F_X}) + MSE(\alpha_P, \alpha_X) \\
\mathcal{L}_{total} = a \mathcal{L}_{cls} + b \mathcal{L}_{dis}
 \end{aligned}
\end{equation}
where $y$ and $\hat{y}$ are the true and predicted intents, $Z_{F_X}$ and $Z_{F_P}$ are the frame level representations of the encoders $f_{ae}$ and $f_P$, $\alpha_X$ and $\alpha_P$ is the attention weights of the student and teacher models. $a$ and $b$ are the MTL weights.

\begin{table}[]
\begin{tabular}{@{}llll@{}}
\toprule
 & \textbf{Train} & \textbf{Validation} & \textbf{Test} \\ \midrule
\textbf{SLURP} & 49943(39.7) & 8561(6.8) & 12951(10.1) \\
\textbf{STOP} & 120906(116.5) & 33385(31.6) & 75510(69.9) \\ \bottomrule
\end{tabular}
\caption{No of utterance(hours) of SLURP and STOP dataset}
\label{tab:dataset-stat}
\vspace{-7mm}
\end{table}

\section{Experiments}

\subsection{Dataset}
We evaluate the proposed methods on two publicly available English SLU datasets, SLURP \cite{bastianelli2020slurp}and STOP \cite{tomasello2023stop}. SLURP is a single-turn user conversation with a home assistant with higher lexical and semantic diversity than most publicly available SLU datasets. STOP is a large dataset that contains utterance-semantic parse pairs. The number of intents is 60 and 64 for SLURP and STOP datasets, respectively. The number of utterances and no of hours for train/validation/test splits of both datasets are provided in Table \ref{tab:dataset-stat}. The main experiments and ablation studies are conducted on the SLURP dataset. We present the final results comparing the proposed method with the baseline on both SLURP and STOP datasets. Both datasets' speech samples are cropped or padded to 5 seconds. 
    
\subsection{Baseline Models}
We used the whisper-based S2I model proposed in \cite{https://doi.org/10.48550/arxiv.2212.13015} as one of the baselines (hereafter whisper baseline) that uses a pretrained whisper(base.en) \cite{radford2022robust} model's encoder as the acoustic encoder, followed by Self-Attention pooling and linear layer for intent prediction. A prosody-based baseline \cite{wei2022neural}(hereafter local concat baseline) was also used for comparison that concatenates prosodic features with the acoustic encoder's 512-dimension representations at each time frame, which is passed through a two-layered LSTM \cite{hochreiter1997long} network with 256 hidden dimension and linear layer for intent classification. 

\subsection{Prosody Attention and Distillation}
The acoustic encoder $f_{ae}$ used in both prosody-attention and prosody-distillation models is the pretrained whisper(base.en) \cite{radford2022robust} model's encoder. The prosody encoder $f_P$ used as the teacher model in prosody-distillation is a sequence of three 1-dimensional Convolutional neural network(CNN) layers with GELU \cite{hendrycks2016gaussian} activation. The CNN layers have a kernel size of 5, with a stride of 1 and 'same' padding to maintain the same frame length. The 6-channel input prosodic features are encoded to a channel dimension of 512 with the CNN layers. The weights $a$ and $b$ for the MTL were initialized randomly for every training step following \cite{https://doi.org/10.48550/arxiv.2111.10603}.

\subsection{Training Setup}
As the whisper model's inputs are Mel Spectrograms, an 80-channel log Mel spectrogram with 25 millisecond windows with a stride of 10 milliseconds is computed for all the speech samples. All the experiments are implemented using Pytorch Lightning and OpenAI's whisper model pretrained checkpoints \cite{radford2022robust}. Adam \cite{kingma2014adam} optimizer was used with $lr=1e^{-5}$ for fine-tuning the pretrained whisper encoder and $lr=1e^{-3}$ for other layers. We used a batch size of 64 and trained for 20 epochs with early stopping patience of 10 epochs. The best checkpoint is saved based on the validation accuracy. We report the model's accuracy score and macro F1 for all the experiments on the test set. The metrics are the mean of three runs with randomly initialized model parameters. All the experiments were performed on an NVIDIA A100 GPU.

\section{Results and Discussion}

\subsection{Benchmark Results}

\begin{table}[]
\centering
\begin{tabular}{@{}lll@{}}
\toprule
\textbf{Method} & \textbf{Accuracy} & \textbf{MF1} \\ \midrule
whisper baseline \cite{https://doi.org/10.48550/arxiv.2212.13015} & 0.6807 & 0.6202 \\
prosody local-concat baseline \cite{https://doi.org/10.48550/arxiv.2205.05590} & 0.6823 & 0.6255 \\ \midrule
\cite{https://doi.org/10.48550/arxiv.2212.13015} with prosody-attention & 0.6887 & 0.6282 \\
\cite{https://doi.org/10.48550/arxiv.2205.05590} with prosody-attention  & 0.6955 & 0.6472 \\
prosody-distillation & \textbf{0.7626} & \textbf{0.7192} \\ \bottomrule
\end{tabular}
\caption{Mean Accuracy and Macro F1 scores of baselines and proposed methods on the SLURP dataset of 3 different runs.}
\label{tab:slurp-results}
\vspace{-6mm}
\end{table}

Table \ref{tab:slurp-results} shows the accuracy and Macro F1 scores for the baselines and proposed methods on the SLURP dataset. We can observe that the prosody local-concat baseline, which concatenated prosodic features with the acoustic encoder's representations, improves the intent classification accuracy by 0.16\%. This shows that prosodic features can help in improving intent classification performance. However, the improvement is minimal with the prosody concatenation method. Analyzing the prosody concat model's learned weights, we found that the model weights for the acoustic encoder representations were five times those for the prosodic features, so the contribution of prosodic features was limited. We added prosody-attention on both the whisper baseline and local concat baseline, improving the test metrics of both baselines by 0.8\% and 1.32\%, respectively. The attention map generated by prosodic features could help identify the frames of the utterance, which contributes more to intent classification than regular self-attention pooling. The prosody-distillation method improves accuracy by a huge margin of 8.19\% and 8.03\% on the whisper and local concat baselines, respectively. 

\begin{table}[htb]
\begin{tabular}{@{}llll@{}}
\toprule
\textbf{Dataset} & \textbf{Method} & \textbf{Accuracy} & \textbf{MF1} \\ \midrule
\textbf{SLURP} & whisper baseline \cite{https://doi.org/10.48550/arxiv.2212.13015} & 0.6807 & 0.6302 \\
\textbf{} & local concat baseline \cite{https://doi.org/10.48550/arxiv.2205.05590} & 0.6823 & 0.6255 \\
\textbf{} & prosody-distillation & \textbf{0.7626} & \textbf{0.7192} \\
\midrule
\textbf{STOP} & whisper baseline \cite{https://doi.org/10.48550/arxiv.2212.13015} & 0.884 & 0.7316 \\
\textbf{} & local concat baseline \cite{https://doi.org/10.48550/arxiv.2205.05590} &  0.889 &  0.7352\\
\textbf{} & prosody-distillation & \textbf{0.917} & \textbf{0.7904} \\ \bottomrule
\end{tabular}
\caption{Mean Accuracy and Macro F1 scores of baselines and proposed methods on SLURP and STOP datasets of 3 different runs.}
\label{tab:final-results}
\vspace{-5mm}
\end{table}

Table \ref{tab:final-results} shows the accuracy and Macro F1 scores of the baselines and proposed prosody-distillation method on SLURP and STOP datasets. The proposed method outperforms both baselines on both datasets without needing implicit prosodic features during inference. This shows the robustness of the proposed method across datasets, and the prosody-distillation method uses prosodic information better than previous methods.

\subsection{Ablation Study}

\begin{table}[]
\begin{tabular}{@{}lll@{}}
\toprule
\textbf{Study} & \multicolumn{1}{c}{\textbf{Method}} & \multicolumn{1}{c}{\textbf{Accuracy}} \\ \midrule
\textbf{S1 : Distillation Layer} & global dist & 0.7107 \\ 
\textbf{} & frame-level dist & \textbf{0.7626} \\ \midrule
\textbf{S2 : Distillation Type} & w/o attn dist & 0.7512 \\
\textbf{} & w/o feature dist & 0.7243 \\ 
\textbf{} & with both & \textbf{0.7626} \\\midrule
\textbf{S3 : Pretraining} & no pretraining & 0.7594 \\
\textbf{} & pretraining & \textbf{0.7626} \\ \midrule
\textbf{S4 : MTL} & a=1, b=1 & 0.7594 \\
\textbf{} & a=1, b=0.1 & 0.7474 \\
\textbf{} & a=0.1, b=1 & 0.7624 \\ 
\textbf{} & a=rand, b=rand \cite{https://doi.org/10.48550/arxiv.2111.10603} & \textbf{0.7626} \\ \midrule
\textbf{S5 : Prosody features} & w/o pitch & 0.7492 \\
\textbf{} & w/o energy & 0.7513 \\
\textbf{} & with both & \textbf{0.7626} \\ \bottomrule
\end{tabular}
\caption{Mean Accuracy scores for ablation studies(S1-S5) on SLURP dataset of 3 different runs.}
\label{tab:ablation-study}
\vspace{-5mm}
\end{table}

In Table \ref{tab:ablation-study}, we present the results of a few ablation studies (S1-S5) that we conducted to find the performance of the proposed method prosody-distillation with different hyperparameters and training methods. All the ablation studies were conducted on the SLURP dataset, and we report the mean accuracy scores of 3 different runs. From S1, we find that feature distillation at a frame level features $z_F$ instead of global utterances level features $z_U$ gave a better performance as prosodic information at different frames is essential to identify the intention. S2 shows that prosody attention and feature distillation are essential, and removing one degrades the model's performance. S3 compares the performance of the prosody-distillation method when the teacher prosody model is pretrained and when the teacher prosody model is trained along with the student model. As expected, pretraining the teacher prosody model gave the best result. When training them together, both attention and features distillation targets will be random initially and keep changing as the teacher model's parameters change. 

In S4, we compare the influence of different weights for the MTL with the weights given in the table. We observed that giving more weight to the distillation loss gave better results than giving more weight to the classification loss or weighing them the same. This suggests the distillation method's importance in learning prosodic information from the teacher model. Moreover, weighting both losses randomly at each iteration slightly improved performance. Finally, in S5, we aim to study the impact of both pitch and energy prosodic features. We trained the prosody-distillation method without pitch and energy features. We can observe that removing pitch features affected the method's performance more than removing energy features and is similar to the result obtained in \cite{https://doi.org/10.48550/arxiv.2205.05590}. So both pitch and energy features are essential for intent classification, but the contribution of the pitch features is more than the energy features.

\subsection{Attention Maps}
\begin{figure}
\centering
  \includegraphics[width=\linewidth]{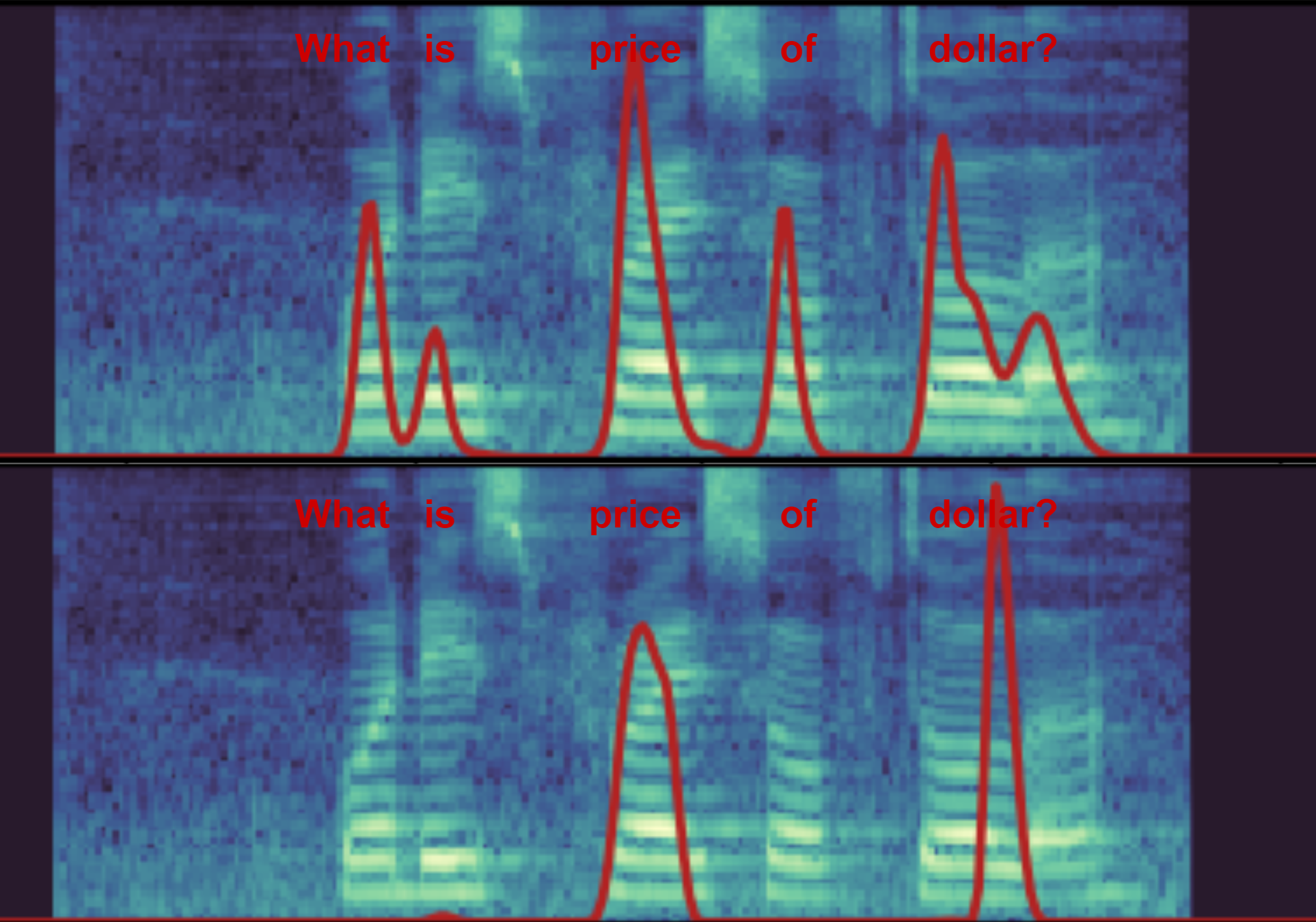}
  \caption{Attention maps of prosody-attention(upper) and prosody-distillation(lower) on a speech utterance from SLURP dataset}
  \label{fig:attention_map2}
  \vspace{-5mm}
\end{figure}

Figure \ref{fig:attention_map2} shows the attention map of the trained prosody-attention and prosody-distillation model. As we can observe, the prosody-attention model ignores all the unvoiced segments and gives maximum attention to the "What", "price", and "dollar" parts of the utterance. The prosody-distillation model gives maximum attention to the word "dollar" and "price" which is more relevant for intent classification and ignores all other words and unvoiced segments. 


\section{Conclusions}
In this work, we propose a new method called prosody-distillation for improving the Speech to Intent performance by using prosodic information without the need for implicit prosodic features. The proposed method performs the knowledge distillation of prosodic attention and prosodic features from a pretrained teacher prosody model. We perform experiments on two publicly available SLU datasets and show that our proposed method improves the performance by huge margins on both datasets. For future work, we want to use prosody-distillation for entity extraction and slot-filling tasks. Also, use other prosodic features and speaker information to improve SLU performance.

\bibliographystyle{IEEEtran}
\bibliography{mybib}

\begin{thebibliography}{10}
\providecommand{\url}[1]{#1}
\csname url@samestyle\endcsname
\providecommand{\newblock}{\relax}
\providecommand{\bibinfo}[2]{#2}
\providecommand{\BIBentrySTDinterwordspacing}{\spaceskip=0pt\relax}
\providecommand{\BIBentryALTinterwordstretchfactor}{4}
\providecommand{\BIBentryALTinterwordspacing}{\spaceskip=\fontdimen2\font plus
\BIBentryALTinterwordstretchfactor\fontdimen3\font minus
  \fontdimen4\font\relax}
\providecommand{\BIBforeignlanguage}[2]{{%
\expandafter\ifx\csname l@#1\endcsname\relax
\typeout{** WARNING: IEEEtran.bst: No hyphenation pattern has been}%
\typeout{** loaded for the language `#1'. Using the pattern for}%
\typeout{** the default language instead.}%
\else
\language=\csname l@#1\endcsname
\fi
#2}}
\providecommand{\BIBdecl}{\relax}
\BIBdecl

\bibitem{tran2020neural}
T.~Tran, \emph{Neural models for integrating prosody in spoken language
  understanding}.\hskip 1em plus 0.5em minus 0.4em\relax University of
  Washington, 2020.

\bibitem{chen2018spoken}
Y.-P. Chen, R.~Price, and S.~Bangalore, ``Spoken language understanding without
  speech recognition,'' in \emph{2018 IEEE International Conference on
  Acoustics, Speech and Signal Processing (ICASSP)}.\hskip 1em plus 0.5em minus
  0.4em\relax IEEE, 2018, pp. 6189--6193.

\bibitem{lugosch2019speech}
L.~Lugosch, M.~Ravanelli, P.~Ignoto, V.~S. Tomar, and Y.~Bengio, ``Speech model
  pre-training for end-to-end spoken language understanding,'' \emph{arXiv
  preprint arXiv:1904.03670}, 2019.

\bibitem{jiang2021knowledge}
Y.~Jiang, B.~Sharma, M.~Madhavi, and H.~Li, ``Knowledge distillation from bert
  transformer to speech transformer for intent classification,'' \emph{arXiv
  preprint arXiv:2108.02598}, 2021.

\bibitem{sunder2022tokenwise}
V.~Sunder, E.~Fosler-Lussier, S.~Thomas, H.-K.~J. Kuo, and B.~Kingsbury,
  ``Tokenwise contrastive pretraining for finer speech-to-bert alignment in
  end-to-end speech-to-intent systems,'' \emph{arXiv preprint
  arXiv:2204.05188}, 2022.

\bibitem{huang2020leveraging}
Y.~Huang, H.-K. Kuo, S.~Thomas, Z.~Kons, K.~Audhkhasi, B.~Kingsbury, R.~Hoory,
  and M.~Picheny, ``Leveraging unpaired text data for training end-to-end
  speech-to-intent systems,'' in \emph{ICASSP 2020-2020 IEEE International
  Conference on Acoustics, Speech and Signal Processing (ICASSP)}.\hskip 1em
  plus 0.5em minus 0.4em\relax IEEE, 2020, pp. 7984--7988.

\bibitem{wallbridge2021s}
S.~Wallbridge, P.~Bell, and C.~Lai, ``It's not what you said, it's how you said
  it: discriminative perception of speech as a multichannel communication
  system,'' \emph{arXiv preprint arXiv:2105.00260}, 2021.

\bibitem{dahan2015prosody}
D.~Dahan, ``Prosody and language comprehension,'' \emph{Wiley Interdisciplinary
  Reviews: Cognitive Science}, vol.~6, no.~5, pp. 441--452, 2015.

\bibitem{https://doi.org/10.48550/arxiv.2212.13015}
\BIBentryALTinterwordspacing
S.~Rajaa, S.~Dalmia, and K.~Nethil, ``Skit-s2i: An indian accented speech to
  intent dataset,'' 2022. [Online]. Available:
  \url{https://arxiv.org/abs/2212.13015}
\BIBentrySTDinterwordspacing

\bibitem{wei2022neural}
K.~Wei, D.~Knox, M.~Radfar, T.~Tran, M.~M{\"u}ller, G.~P. Strimel, N.~Susanj,
  A.~Mouchtaris, and M.~Omologo, ``A neural prosody encoder for end-to-end
  dialogue act classification,'' in \emph{ICASSP 2022-2022 IEEE International
  Conference on Acoustics, Speech and Signal Processing (ICASSP)}.\hskip 1em
  plus 0.5em minus 0.4em\relax IEEE, 2022, pp. 7047--7051.

\bibitem{bernard2023shennong}
M.~Bernard, M.~Poli, J.~Karadayi, and E.~Dupoux, ``Shennong: a python toolbox
  for audio speech features extraction,'' \emph{Behavior Research Methods}, pp.
  1--13, 2023.

\bibitem{safari2020self}
P.~Safari, M.~India, and J.~Hernando, ``Self-attention encoding and pooling for
  speaker recognition,'' \emph{arXiv preprint arXiv:2008.01077}, 2020.

\bibitem{bastianelli2020slurp}
E.~Bastianelli, A.~Vanzo, P.~Swietojanski, and V.~Rieser, ``Slurp: A spoken
  language understanding resource package,'' \emph{arXiv preprint
  arXiv:2011.13205}, 2020.

\bibitem{tomasello2023stop}
P.~Tomasello, A.~Shrivastava, D.~Lazar, P.-C. Hsu, D.~Le, A.~Sagar, A.~Elkahky,
  J.~Copet, W.-N. Hsu, Y.~Adi \emph{et~al.}, ``Stop: A dataset for spoken task
  oriented semantic parsing,'' in \emph{2022 IEEE Spoken Language Technology
  Workshop (SLT)}.\hskip 1em plus 0.5em minus 0.4em\relax IEEE, 2023, pp.
  991--998.

\bibitem{radford2022robust}
A.~Radford, J.~W. Kim, T.~Xu, G.~Brockman, C.~McLeavey, and I.~Sutskever,
  ``Robust speech recognition via large-scale weak supervision,'' \emph{arXiv
  preprint arXiv:2212.04356}, 2022.

\bibitem{hochreiter1997long}
S.~Hochreiter and J.~Schmidhuber, ``Long short-term memory,'' \emph{Neural
  computation}, vol.~9, no.~8, pp. 1735--1780, 1997.

\bibitem{hendrycks2016gaussian}
D.~Hendrycks and K.~Gimpel, ``Gaussian error linear units (gelus),''
  \emph{arXiv preprint arXiv:1606.08415}, 2016.

\bibitem{https://doi.org/10.48550/arxiv.2111.10603}
\BIBentryALTinterwordspacing
B.~Lin, F.~Ye, Y.~Zhang, and I.~W. Tsang, ``Reasonable effectiveness of random
  weighting: A litmus test for multi-task learning,'' 2021. [Online].
  Available: \url{https://arxiv.org/abs/2111.10603}
\BIBentrySTDinterwordspacing

\bibitem{kingma2014adam}
D.~P. Kingma and J.~Ba, ``Adam: A method for stochastic optimization,''
  \emph{arXiv preprint arXiv:1412.6980}, 2014.

\bibitem{https://doi.org/10.48550/arxiv.2205.05590}
\BIBentryALTinterwordspacing
K.~Wei, D.~Knox, M.~Radfar, T.~Tran, M.~Muller, G.~P. Strimel, N.~Susanj,
  A.~Mouchtaris, and M.~Omologo, ``A neural prosody encoder for end-ro-end
  dialogue act classification,'' 2022. [Online]. Available:
  \url{https://arxiv.org/abs/2205.05590}
\BIBentrySTDinterwordspacing

\end{thebibliography}

\end{document}